\documentclass{sigchi}

\CopyrightYear{2021}
\setcopyright{acmlicensed}
\doi{https://doi.org/10.1145/3313831.XXXXXXX}
\isbn{978-1-4503-6708-0/20/04}
\conferenceinfo{HRI'21,Workshop "Exploring Applications for Autonomous Non-Verbal Human-Robot Interactions"}{March  8, 2021}

\usepackage{balance}       
\usepackage{graphics}      
\usepackage[T1]{fontenc}   
\usepackage{txfonts}
\usepackage{mathptmx}
\usepackage[pdflang={en-US},pdftex]{hyperref}
\usepackage{color}
\usepackage{booktabs}
\usepackage{textcomp}

\usepackage{microtype}        
\usepackage{ccicons}          

\usepackage{todonotes}

\def\plaintitle{A proxemics game between festival visitors and an industrial robot}

\def\emptyauthor{}
\def\plainkeywords{non-verbal human-robot interaction; cobots; data collection experiment; qualitative study}

\makeatletter
\def\url@leostyle{%
  \@ifundefined{selectfont}{
    \def\UrlFont{\sf}
  }{
    \def\UrlFont{\small\bf\ttfamily}
  }}
\makeatother
\def\pprw{8.5in}
\def\pprh{11in}

\definecolor{linkColor}{RGB}{6,125,233}
\hypersetup{%
  pdftitle={\plaintitle},
  pdfauthor={\emptyauthor},
  pdfkeywords={\plainkeywords},
  pdfdisplaydoctitle=true, 
  bookmarksnumbered,
  pdfstartview={FitH},
  colorlinks,
  citecolor=black,
  filecolor=black,
  linkcolor=black,
  urlcolor=linkColor,
  breaklinks=true,
  hypertexnames=false
}

\begin{document}

\title{\plaintitle}

\numberofauthors{3}
\author{%
  \alignauthor{Brigitte Krenn\\Stephanie Gross\\
    \affaddr{Austrian Research Institute for Artificial Intelligence}\\
    \affaddr{Vienna, Austria}\\
    \email{firstname.lastname@ofai.at}}\\
  \alignauthor{Bernhard Dieber\\Horst Pichler\\
    \affaddr{Joanneum Research - Robotics}\\
    \affaddr{Klagenfurt, Austria}\\
    \email{firstname.lastname@joanneum.at }}\\
  \alignauthor{Kathrin Meyer\\
    \affaddr{LIT Robopsychology Lab, Johannes Kepler University}\\
    \affaddr{Linz, Austria}\\
    \email{kathrin.meyer@jku.at}}\\
}

\maketitle

\begin{abstract}
With increased applications of collaborative robots (cobots) in industrial workplaces, behavioural effects of human-cobot interactions need to be further investigated. This is of particular importance as nonverbal behaviours of collaboration partners in human-robot teams significantly influence the experience of the human interaction partners and the success of the collaborative task. 
During  the Ars Electronica 2020 Festival for Art, Technology \& Society in Linz, we invited visitors to exploratively interact with an industrial robot, exhibiting restricted interaction capabilities: extending and retracting its arm, depending on the movements of the volunteer. The movements of the arm were pre-programmed and telecontrolled for safety reasons (which was not obvious to the participants). We recorded video data of these interactions and investigated general nonverbal behaviours of the humans interacting with the robot, as well as nonverbal behaviours of people in the audience. Our results showed that people were more interested in exploring the robot's action and perception capabilities than just reproducing the interaction game as introduced by the instructors. We also found that the majority of participants interacting with the robot approached it up to a distance which would be perceived as threatening or intimidating, if it were a human interaction partner. Regarding bystanders, we found examples where people made movements as if trying out variants of the current participant's behaviour.
\end{abstract}




\keywords{\plainkeywords}

\printccsdesc

\section{Introduction}


Industrial robotics is a field of increasing relevance. Sensor systems of robots are continuously improved, opening new possibilities to allow safe collaborations with humans. In interactions with collaborative robots (cobots), multimodal signals from the interaction partner (human or robot) need to be identified and interpreted, giving insights in their respective intentions. These communicative aspects are of great importance to facilitate successful joint task completion and to foster user acceptance and optimize the user’s experience \cite{Saunderson2019}, and define  requirements for robot engineers to design trustworthy and transparent robotic AI \cite{EUHLEG2019}.

\begin{figure}
    \centering
    \includegraphics[width=.25\textwidth]{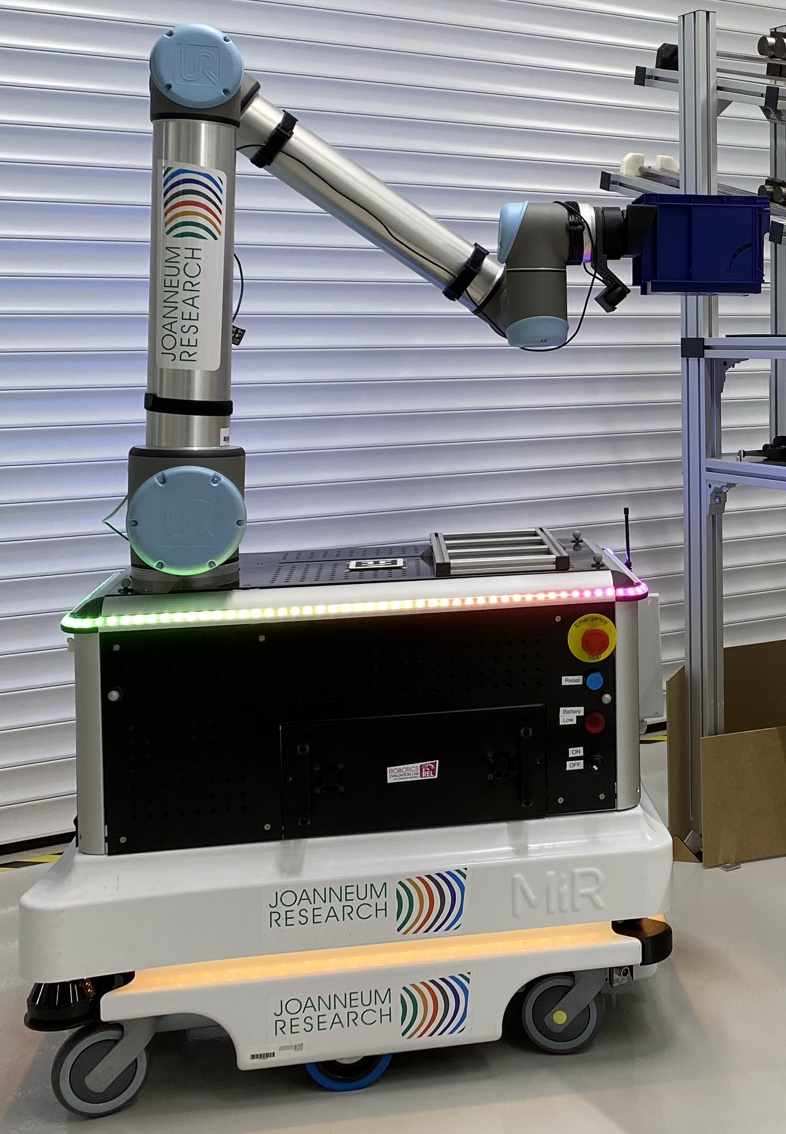}
    \caption{The CHIMERA robot, a combination of UR10 robot arm and MiR100 mobile platform. Here CHIMERA is depicted in the lab carrying a box.}
    \label{fig:chimera}
\end{figure}

In this paper, we present a data collection experiment where we brought together naive users with CHIMERA\footnote{\url{https://www.joanneum.at/en/robotics/infrastructure/mobile-manipulation}} (Figure \ref{fig:chimera}), a custom made sensitive robot for industrial and academic research. CHIMERA combines an UR10 robot arm and a MiR-100 mobile platform to an overall versatile system. UR10 is a  collaborative industrial robot arm with long reach capability, 10 stands for 10kg payload.\footnote{\url{https://www.universal-robots.com/products/ur10-robot/}} MiR-100 is a mobile robot platform, designed with a focus on safety, as it is able to navigate around obstacles and people, or to stop if the system finds no bypass.\footnote{\url{https://www.mobile-industrial-robots.com/en/solutions/robots/mir100/}} This characteristics make the two robotic components well suited for collaborative robotics applications in industry, and CHIMERA a convenient robot for researching  close encounters of humans with cobots. 

The explorative study is part of a larger research project "CoBot Studio -- Crossing Realities for Mutual Understanding in Human-Robot Teams"\footnote{\url{https://www.cobotstudio.org/}} which researches human-robot interaction from a collaborative point of view where the mutual ability of human and robot to understand each other's signals is core to the successful completion of joint tasks, and also plays a major role for human acceptance of and trust in the robot as a workmate. In the project, we study close collaborations of humans with cobots employing both Virtual Reality (VR) scenarios and real world encounters. The advantage of VR is that it allows for very close physical proximity between human and robot, and for robot action speeds that would be impossible near humans in reality due to safety and security regulations. However, to ensure comparability with real world interactions, both environments are employed in user studies where CHIMERA is used as a comparatively safe and agile cobot. 

The goal of the study presented in this paper was to collect qualitative data on how naive users react to, interact with and interpret an apparently interactive industrial robot. The cobot was staged at the Ars Electronica 2020 Festival for Art, Technology \& Society,\footnote{\url{https://ars.electronica.art/keplersgardens/en/}} which  provided us with a constant flow of  visitors. Human and robot were to play a kind of proxemics game where the robot in a one-to-one encounter with the human tries to keep its arm length distance from the human. The robot arm could move into three positions: center, 
forward, and backward. 
The setting is derived from psychology research on proxemics, i.e., how people, typically subconsciously, organize the space between themselves and their interaction partners \cite{hall1974handbook,hall1968proxemics}. 

As CHIMERA is a custom made robot and its actions were programmed particularly for the presented data collection experiment, we can be very sure that the visitors had not seen or interacted with the robot before. 
This was exactly what we aimed at: collect video data from first encounters of naive users with a cobot making use of very limited and strictly non-verbal communicative signals (movements of the platform and the Tool Center Point of the robotic arm). Having manually analysed the video recordings from one day of human-robot interactions, capturing both the robot's direct interaction partner and  bystanders, we can summarize our main findings as follows: (i) there was a remarkable lack of spatial distance from the human interaction partner towards the robot, (ii) people showed a tendency for explorative play with the robot, (iii) some people assumed that there is a motion sensor below the gripper, (iv) some of the onlookers moved as if trying out variants of the current participant's behaviour. 


In the next sections, we first present related work. Subsequently, the technical realisation of the cobot behaviour is described as well as the scenario and context where the data collection took place. Qualitative analysis and findings are then presented. This is followed by a conclusion and discussion of future directions.  

\section{Related Work}

The distance between two agents during an interaction is a fundamental principle of social interaction and known as proxemics. Proxemics was first described by anthropologist Eduard Hall, who emphasized the impact of the use of space, i.e. proxemic behavior, on interpersonal communication \cite{hall1974handbook}.  Hall defined four primary proximity zones \cite{Hall1966}:\footnote{The units are translated from inch/feet in the original work to cm/m. One should also keep in mind that the borders between the zones are more guidelines than strict numbers.}
\begin{itemize}
    \item intimate (contact - 0.46m): two agents can touch, smell, talk in whisper etc., but cannot see each other well.
    \item personal (0.46m - 1.22m): personal distance corresponds to personal space, where discomfort can be felt if that space is penetrated by someone whom the individual is unfamiliar with and it is possible to touch each other by reaching.
    \item social (1.22-3.66m): social distances are used in formal business purposes, e.g. across a desk with the need to speak louder.
    \item public (3.66m and beyond): public distance is the distance kept from public figures, facial expressions are difficult to see, a louder voice is needed and bodily movements need to be exaggerated.
\end{itemize}

Proxemics has been shown to apply to human-robot interactions (see e.g. \cite{Saunderson2019}), which is consistent with several studies demonstrating that individuals respond socially to computers \cite{Nass2000}. In human-robot interaction, the robot’s effective use of proxemic behaviors can be crucial for the engagement with humans as it allows for intuitive interaction between humans and robots \cite{Saunderson2019}.

In a study on comfortable approach distances, Walters et al.  \cite{Walters2005} investigated both the robot approach perspective and the human approach perspective, using the PeopleBot telepresence robot. They found that 60\% of the participants were comfortable with a social distance compatible with human-human interaction distances. However, 40\% actually approached the robot up to a distance which would be perceived as either threatening or intimate, depending if it involves strangers or close friends. When investigating human personality traits, the authors found that timidity and nervousness increased the comfortable approach distance, whereas it was decreased by traits such as proactiveness. 
In a subsequent long-term study, Walters et al. \cite{Walters2011} showed reductions in comfortable human-robot approach distances in the first one to two interaction sessions. For the remaining sessions, approach distance preferences remained relatively steady.

Koay et al. \cite{Koay2007} found that humans allowed robots to approach more closely during physical interactions (e.g. hand-over tasks) than during verbal interactions or no interaction conditions (e.g. when the robot moves through the same room). In their study, participants also exhibited a strong tendency to allow the robot with mechanoid appearance to approach closer during their first encounter than the robot with humanoid appearance. However, results showed that this tendency faded away as the participants habituated to the robots. In a subsequent study, Koay et al. \cite{Koay2014} found evidence that participants preferred to be approached from the front versus the side. This preference increased with closer distances.

Shi et al. \cite{Shi2008} investigated perceived safety levels associated with distance and velocity, using a Segway robotic platform. Results of their study showed that distance itself did not effect participants. However, for the fast velocity condition participants felt slightly unsafe and some even moved away from the robot's path as it approached them. 

In our study, the robot was placed at a certain location in the room. After a participant had approached the robot, the interaction started. The robot remained in its position during the interaction for safety reasons, while the robot arm moved back and forth depending on the human's movements.

\section{Robot Technical Set-up}
\label{sec_tech}

CHIMERA (Figure \ref{fig:chimera}) is a combination of two commercial robots: MiR-100 (the mobile base) and  Universal Robots UR-10 (the robot arm). They are integrated into CHIMERA's (henceforth the robot's) internal network. Mobile base and arm have their own programming methods supplied by the respective vendors but can be interconnected for a limited set of functions (e.g., triggering a program on the robot arm as part of a mission of the robot base). However, running on CHIMERA's internal industrial PC, there is a software stack that enables joint programs of both MiR and UR when the robot performs autonomous operations.

For safety and security reasons in a public space, however, both, mobile platform and robot arm were human operated by two researchers from our team (which was not obvious to the participants). The robot arm was pre-programmed with a reactive application that responds to a two-button remote. Using this, the robot arm could be moved into a center (idle), a forward (approach) or a rear (retract) position. The mobile base first recorded a map of the surrounding area. Based on that, missions were programmed which guided the robot through the exhibition area. Also direct manual remote control of the mobile base was used where appropriate.

\section{Data Collection Scenario}
\label{sec_dc}

\begin{figure}
\centering
  \includegraphics[width=1\columnwidth]{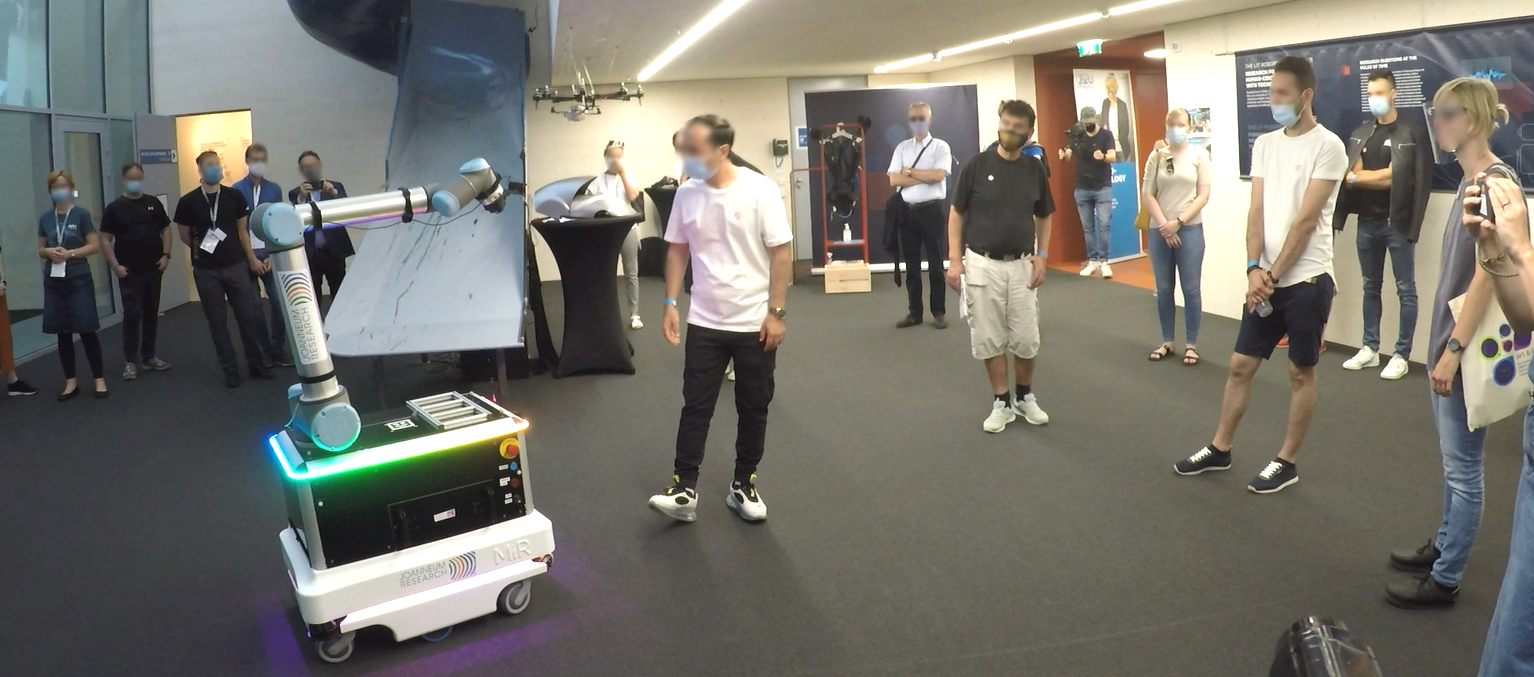}
  \caption{Location at Ars Electronica 2020 where human-robot encounters took place. View from statically mounted Gopro.}~\label{fig:figure1}
\end{figure}

The users were visitors to Kepler's Gardens at the Ars Electronica Festival 2020. Figure \ref{fig:figure1} shows the exhibition space of the LIT Robopsychology Lab\footnote{\url{https://www.jku.at/en/lit-robopsychology-lab/}} where the human-robot encounters took place. The room was big enough (105 m\textsuperscript{2}) for the robot platform to move and the arm to extend to its maximum length, and at the same time give the human plenty of space to maneuver. Due to COVID-19 restrictions, visitors had to book slots in advance and came in groups of approximately 20 people or less. Visiting groups were introduced to the robot by one of the researchers present. They were told that the robot reacts to proximity, i.e., when a person stands in front of the robot and s/he gives way to the robot, the arm will move forward in the human's direction. If s/he leans forward towards the robot, the arm will move back. In other words, the robot wants to keep its "arm length" distance to the human. The visitors were also told that they should make sure that they kept their heads away from the robot arm so that they cannot be harmed. The interaction was then demonstrated to the visitors by the researcher in charge and they were invited to try it out themselves, one by one. For data collection, video recordings were taken from the visits during 12 September 2020. Videos were taken with (i) a statically mounted GoPro directed at the robot and the human interaction partner from the right side, and covering the background in a wide angle from a distance of approximately 5 meters; (ii) a camera mounted on the end effector of the robot arm providing a frontal view on the human interaction partner and the scene in the background whenever the human was not too close to the robot's camera; (iii) in addition, some of the scenes were also captured by videos taken by one or two members of the research team moving around with mobile phone cameras. This provided us with plenty of material for studying the behaviours of the human directly interacting with the robot and gave us impressions of bystanders' reactions. 

\section{Qualitative Analysis and Findings}
\label{sec_ana}



The videos were sifted with respect to what is the non-verbal behaviour of the human interacting with the robot, and what do people in the audience do? First of all, we analysed the recordings from the Gopro and the mobile phone cameras. In a first round, we were interested in the behaviours of the robot's interaction partners. Aspects of interest were: does the human more or less copy the behaviour of the presenter; to which extent does the human reproduce the proxemics game, i.e., does the human give way or take space so that the robot arm can move forward towards the human or backs off; what kinds of explorative behaviour does the human show; what are the spatial distances the human maintains towards the robot. In a second round, we concentrated on the bystanders in our video material. As regards the audience, we did not have any detailed assumptions, we were generally interested in typical behaviours that would occur. In a third round, we looked into the videos from the camera mounted at the robot's end effector, to get a first person perspective from the robot's side. These data are particularly interesting, as they vividly depict how close to the robot human interaction partners tend to come, either with their faces or at least their hands.  

\subsection{Behaviours of Humans Interacting with the Robot}

\paragraph{Human reproduction of game play} Interestingly, there were only a few visitors who just interacted with the robot in the way we had demonstrated the interaction. There was also a small number who seemed to be waiting for the robot to make the first step or even more they seemed to expect the robot to guide them through the interaction. If this did not happen, they lost interest. The majority, however, were much more interested in exploring the robot. 

\paragraph{Human explorative behaviours} Quite a number of people were much more interested in exploring the robot than just interacting in the demonstrated manner. People seemed to suspect that the end effector had some kind of sensor as people tended to hold their hands very close below the end effector and tried to make the robot follow their hand movements. As if the end effector were the robot's snout. This may possibly be due to the slight up and down movements of the end effector when moving the robot arm back and forth, which were a side effect of keeping the camera in forward orientation. This made the end part of the robot arm the most vivid part of the robot during the interaction. Besides, it was also the smallest part of the robot and fitted very well with the size of the human palm. Other frequent human behaviours were  gesturing in front of the camera mounted on the end effector, or getting very close to it with one's face. For the former, as if the humans wanted to test if the robot's arm would follow their moving hand or index finger. Just like in gaze following games where one person gets very close to the face of the other person and makes the other person's eyes follow their moving hand or index finger. As regards the latter, we had the impression as if the humans wanted to provoke the robot to show a reaction by them massively invading the robot's intimate space. In general, we had the impression that the human interaction partners were seeking for the robot's sensors, typically by gesticulating and coming very close. Sensors were expected (i) at the end of the end effector; (ii) the camera mounted on the end effector was identified; and (iii) the frontal sides (left and right) of the mobile platform were searched for further sensors. 

\paragraph{Proxemics} Even though the visitors were 
instructed to keep at least an arm length distance from the robot, putting one's arm forward and at the same time keeping one's head well away from the robot arm, the volunteers came very close to the robot during interaction. 
Distances ranged from close personal to intimate distance. In this respect our data provide further evidence for Koay et al. and Walters et al., that people during first encounters tend to come rather close to robots \cite{Walters2005} with mechanoid appearance \cite{Koay2007, Koay2014}. Walters et al. \cite{Walters2005} hypothesized that in these very close interactions, humans do not treat the robot as a social entity.

We did not see anybody approaching the robot from behind. This may be due to the fact that the robot had a clear front to rear orientation. It was positioned in the room opposite of the entrance with the arm facing the incoming visitors, and the arm going backward and forward in the proxemics game supported this orientation, too.

Another observation was that when children and adults stepped forward together to interact with the robot, it happened that when one started the interaction with the robot, the other one quite quickly started to push the interacting one aside, in order to try out the interaction by themselves.

\subsection{Behaviours of Bystanders}
In the behaviours of the onlookers, we found the following typicalities: (i) the audience uniformly kept a wide social up to public distance during the presentation by the instructor and while someone else was interacting with the robot, (ii) attention according to head gaze was clearly oriented towards the scene of human-robot interaction. In other words, the bystanders were observing the scene from a distance. (iii) In some cases, we found dry runs of onlookers imitating the person currently interacting with the robot. Moreover, it seemed that people were trying variants of the behaviour the robot's current interaction partner displayed. Such behaviours were displayed in cases where the robot did not seem to respond to the human, even though the human displayed a behaviour similar to the one demonstrated by the instructor. It seemed as if the dry runs were aimed at exploring potentially more successful variants of the  behaviour. It may also be that those individuals who made their dry runs were interested in the interaction but were too shy to step forward and actually interact with the robot.



\section{Conclusion and Future Work}
\label{sec_conc}

We presented a qualitative study comprising a simple proxemics game between a cobot and walk-in visitors of a festival for arts, technology and society. From the data we found that people had a high interest in exploring the robots action and perception capabilities, much more than just reproducing the interaction game as they were told by the instructors. We also found that people who interacted with the robot did this from very close proximity, especially when they seemed to explore the robot. In addition, we found examples of behavioural dry runs by bystanders who seemed to test out variants of the interaction behaviour which was presented by the instructors and did not seem to immediately trigger the expected reaction from the robot by the human currently interacting with the robot. 

Overall, our qualitative study was a first step towards further investigating the interplay of robot communicative signals and human non-verbal behaviours in interactions with cobots. From the findings, we derive, on the one hand, technical requirements for robot action and perception capabilities including the necessity of a cobot (i) to check the attention of its human collaboration partner, in particular, which part of the robot the human looks at or gets close to with their hands or faces; the interpretation of which also requires (ii) a model of proxemics; (iii) to signal to the human where the robot's current focus of attention is, e.g. by change of orientation of the robot arm; (iv) to use movement (approaching, moving away, facing away and turning around) as communicative signals for proxemics and attention to the human; (v) to be able to identify and correctly interpret human gesturing such as beckoning vs. pointing, emblems e.g. stop. 


For autonomous or semi-autonomous robots interacting with humans, perceiving and interpreting human motions correctly and subsequently following some social behaviors is considered as one of the most important capabilities \cite{Yan2014}. Therefore, the findings are very valuable inputs for engineers, as they will add significantly to the robot's transparency towards humans in order to increase the trustworthiness of autonomous robots.  

On the other hand, we derive questions for strands of further experimentation, both qualitative and quantitative which include: studying human behaviour changes in long term interactions with cobots, or
studying human-cobot interaction behaviours for specific industrial interacion scenarios, ideally with industrial workers/trainees. 

Another research question requiring further investigations is to what extent an artifical agent is perceived as a social agent, especially when proximity zones of the artificial agent are frequently violated.

With regards to behavioural effects of humans interacting with cobots, an important aspect of future work is to more closely investigate differences in behaviour and their origin. They might occur due to specific requirements of a certain collaboration task, or be influenced by the coworker's personality and other psychologically motivated aspects.
As regards the former, in some tasks mutual gaze behaviour between the collaboration partners is essential for task success; in others joint attention is heavily directed by which objects are currently manipulated by one or both collaboration partners; posture shifts can be strong indicators for whose turn it is to do the next working step  \cite{Schreitter2014,Gross2017}. As regards the latter, the effects of personality traits \cite{Walters2005}, trust \cite{Sadrfaridpour2016} and habituation \cite{Koay2007}, as well as other socio-emotional aspects \cite{Shi2008} or perceived job security may have significant effects on human behaviours in the interaction with cobots.

\section{Acknowledgments}

This work was supported by the Austrian Research Promotion Agency (FFG), Ideen Lab 4.0 project CoBot Studio (872590), and the Vienna Science and Technology Fund (WWTF), project "Human tutoring of robots in industry" (NXT19-005).

\balance{}

\bibliographystyle{SIGCHI-Reference-Format}
\bibliography{paper}


\begin{thebibliography}{00}


\ifx \showCODEN    \undefined \def \showCODEN     #1{\unskip}     \fi
\ifx \showDOI      \undefined \def \showDOI       #1{{\tt DOI:}\penalty0{#1}\ }
  \fi
\ifx \showISBNx    \undefined \def \showISBNx     #1{\unskip}     \fi
\ifx \showISBNxiii \undefined \def \showISBNxiii  #1{\unskip}     \fi
\ifx \showISSN     \undefined \def \showISSN      #1{\unskip}     \fi
\ifx \showLCCN     \undefined \def \showLCCN      #1{\unskip}     \fi
\ifx \shownote     \undefined \def \shownote      #1{#1}          \fi
\ifx \showarticletitle \undefined \def \showarticletitle #1{#1}   \fi
\ifx \showURL      \undefined \def \showURL       #1{#1}          \fi

\bibitem{Gross2017}
{Stephanie Gross}, {Brigitte Krenn}, {and} {Matthias Scheutz}. 2017.
\newblock \showarticletitle{The reliability of non-verbal cues for situated
  reference resolution and their interplay with language: implications for
  human robot interaction}. In {\em Proceedings of the 19th ACM International
  Conference on Multimodal Interaction}. 189--196.
\newblock


\bibitem{Hall1966}
{Edward~Twitchell Hall}. 1966.
\newblock {\em The hidden dimension}. Vol. 609.
\newblock Garden City, NY: Doubleday.
\newblock


\bibitem{hall1974handbook}
{Edward~Twitchell Hall}. 1974.
\newblock {\em Handbook for proxemic research}.
\newblock Amer Anthropological Assn.
\newblock


\bibitem{hall1968proxemics}
{Edward~T Hall}, {Ray~L Birdwhistell}, {Bernhard Bock}, {Paul Bohannan},
  {A~Richard Diebold~Jr}, {Marshall Durbin}, {Munro~S Edmonson}, {JL Fischer},
  {Dell Hymes}, {Solon~T Kimball}, {and} {others}. 1968.
\newblock \showarticletitle{Proxemics [and comments and replies]}.
\newblock {\em Current anthropology\/} {9}, 2/3 (1968), 83--108.
\newblock


\bibitem{Koay2014}
{Kheng~Lee Koay}, {Dag~Sverre Syrdal}, {Mohammadreza Ashgari-Oskoei},
  {Michael~L Walters}, {and} {Kerstin Dautenhahn}. 2014.
\newblock \showarticletitle{Social roles and baseline proxemic preferences for
  a domestic service robot}.
\newblock {\em International Journal of Social Robotics\/} {6}, 4 (2014),
  469--488.
\newblock


\bibitem{Koay2007}
{Kheng~Lee Koay}, {Dag~Sverre Syrdal}, {Michael~L Walters}, {and} {Kerstin
  Dautenhahn}. 2007.
\newblock \showarticletitle{Living with robots: Investigating the habituation
  effect in participants' preferences during a longitudinal human-robot
  interaction study}. In {\em RO-MAN 2007-The 16th IEEE International Symposium
  on Robot and Human Interactive Communication}. IEEE, 564--569.
\newblock


\bibitem{Nass2000}
{Clifford Nass} {and} {Youngme Moon}. 2000.
\newblock \showarticletitle{Machines and mindlessness: Social responses to
  computers}.
\newblock {\em Journal of social issues\/} {56}, 1 (2000), 81--103.
\newblock


\bibitem{EUHLEG2019}
{EU~High-Level Expert~Group on Artifical~Intelligenc}. 2019.
\newblock \showarticletitle{Ethics Guidelines for Trustworthy AI}.
\newblock  (2019).
\newblock


\bibitem{Sadrfaridpour2016}
{Behzad Sadrfaridpour}, {Hamed Saeidi}, {and} {Yue Wang}. 2016.
\newblock \showarticletitle{An integrated framework for human-robot
  collaborative assembly in hybrid manufacturing cells}. In {\em 2016 IEEE
  international conference on automation science and engineering (CASE)}. IEEE,
  462--467.
\newblock


\bibitem{Saunderson2019}
{Shane Saunderson} {and} {Goldie Nejat}. 2019.
\newblock \showarticletitle{How robots influence humans: A survey of nonverbal
  communication in social human--robot interaction}.
\newblock {\em International Journal of Social Robotics\/} {11}, 4 (2019),
  575--608.
\newblock


\bibitem{Schreitter2014}
{Stephanie Schreitter} {and} {Brigitte Krenn}. 2014.
\newblock \showarticletitle{Exploring inter-and intra-speaker variability in
  multi-modal task descriptions}. In {\em The 23rd IEEE International Symposium
  on Robot and Human Interactive Communication}. IEEE, 43--48.
\newblock


\bibitem{Shi2008}
{Dongqing Shi}, {Emmanuel~G Collins~Jr}, {Brian Goldiez}, {Arturo Donate},
  {Xiuwen Liu}, {and} {Damion Dunlap}. 2008.
\newblock \showarticletitle{Human-aware robot motion planning with velocity
  constraints}. In {\em 2008 International Symposium on Collaborative
  Technologies and Systems}. IEEE, 490--497.
\newblock


\bibitem{Walters2005}
{Michael~L Walters}, {Kerstin Dautenhahn}, {Ren{\'e} Te~Boekhorst}, {Kheng~Lee
  Koay}, {Christina Kaouri}, {Sarah Woods}, {Chrystopher Nehaniv}, {David Lee},
  {and} {Iain Werry}. 2005.
\newblock \showarticletitle{The influence of subjects' personality traits on
  personal spatial zones in a human-robot interaction experiment}. In {\em
  ROMAN 2005. IEEE International Workshop on Robot and Human Interactive
  Communication, 2005.} IEEE, 347--352.
\newblock


\bibitem{Walters2011}
{Michael~L Walters}, {Mohammedreza~A Oskoei}, {Dag~Sverre Syrdal}, {and}
  {Kerstin Dautenhahn}. 2011.
\newblock \showarticletitle{A long-term human-robot proxemic study}. In {\em
  2011 RO-MAN}. IEEE, 137--142.
\newblock


\bibitem{Yan2014}
{Haibin Yan}, {Marcelo Jr}, {and} {Aun-Neow Poo}. 2014.
\newblock \showarticletitle{A Survey on Perception Methods for Humann-Robot
  Interaction in Social Robots}.
\newblock {\em International Journal of Social Robotics\/}  {6} (01 2014).
\newblock
\showDOI{%
\url{http://dx.doi.org/10.1007/s12369-013-0199-6}}


\end{thebibliography}

\end{document}